\documentclass{article}

\usepackage[nonatbib, preprint]{neurips_2023}

\usepackage[utf8]{inputenc} 
\usepackage[T1]{fontenc}    
\usepackage{hyperref}       
\usepackage{url}            
\usepackage{amsfonts}       
\usepackage{nicefrac}       
\usepackage{microtype}      
\usepackage[table]{xcolor}         
\usepackage{wrapfig}

\usepackage{booktabs}       
\usepackage{listings}
\usepackage{tablefootnote}
\usepackage{multirow}
\usepackage{comment}

\newcommand\Tstrut{\rule{0pt}{2.6ex}}         

\usepackage{caption}
\usepackage[frozencache=true, cachedir=minted-cache]{minted}
\usepackage[most,minted]{tcolorbox}

\usepackage{array}
\newcolumntype{P}[1]{>{\centering\arraybackslash}p{#1}}
\definecolor{backcolour}{rgb}{0.95,0.95,0.92}

\tcbset{
listing engine=minted,
    listing only,
    boxrule=1pt,
  minted options={
    fontsize=\scriptsize, 
    breaklines=True,
  },
     }
\title{Can Large Language Models Really Improve by Self-critiquing Their Own Plans?}

\author{%
   Karthik Valmeekam\thanks{Equal Contribution} \\
  School of Computing \& AI \\Arizona State University 
  Tempe.\\
  \texttt{kvalmeek@asu.edu} \\
  \And
     Matthew Marquez$^*$\\
  School of Computing \& AI\\ Arizona State University,
  Tempe.\\
   \texttt{mmarqu22@asu.edu} \\
  \And
     Subbarao Kambhampati\\
  School of Computing \& AI\\ Arizona State University,
  Tempe.\\
  \texttt{rao@asu.edu} 
}

\begin{document}
\maketitle
\begin{abstract}
There have been widespread claims about Large Language Models (LLMs) being able to successfully verify or self-critique their candidate solutions in reasoning problems in an iterative mode. Intrigued by those claims, in this paper we set out to investigate the verification/self-critiquing abilities of large language models in the context of planning. We evaluate a planning system that employs LLMs for both plan generation and verification. We assess the verifier LLM's performance against ground-truth verification, the impact of self-critiquing on plan generation, and the influence of varying feedback levels on system performance. Using GPT-4, a state-of-the-art LLM, for both generation and verification, our findings reveal that self-critiquing appears to diminish plan generation performance, especially when compared to systems with external, sound verifiers and the LLM verifiers in that system produce a notable number of false positives, compromising the system's reliability. Additionally, the nature of feedback, whether binary or detailed, showed minimal impact on plan generation. Collectively, our results cast doubt on the effectiveness of LLMs in a self-critiquing, iterative framework for planning tasks.

\end{abstract}
\section{Introduction}
Large Language Models have rapidly captured the attention of the AI research community with their exceptional natural language completion capabilities. Trained on web-scale language corpora, these models have demonstrated the ability to generate seemingly valuable completions across a wide range of topics. This led to a surge of  interest in determining whether such models were able to perform well on reasoning tasks. Even though initial anecdotal results showed promise, systematic studies revealed their incompetency in reasoning -- be it planning \cite{valmeekam2023planning} or in simple arithmetic or logic \cite{dziri2023faith}. These results questioning the robustness of their reasoning abilities led to researchers exploring ways to improve these systems. Of particular interest to us is the emerging research on self-critiquing, where the LLMs are used to critique their own candidate generations and iterate. The current works \cite{yao2023tree, shinn2023reflexion, weng2022large} exhibit considerable optimism about using LLMs to critique their own candidate generations, especially in an iterative setting where they keep refining their candidate generations. Additionally, the notion that verifying correctness is computationally simpler than generation for reasoning adds to the optimism. However, there are grounds to be skeptical about it as the complexity of a reasoning task in the classical sense should be irrelevant to models like LLMs that do approximate retrieval.

Intrigued by the prevailing optimism, in this paper, we set out to systematically investigate the effectiveness of using LLMs to critique their own generations in the context of planning. We look at the simplest class of planning problems, the goal-directed deterministic planning problems colloquially referred to as classical planning problems. Our methodology employs a planning system that utilizes the same LLM for both generation and verification, which we term the LLM+LLM system in an iterative setting. Within this setting, the generator LLM continuously produces candidate plans, drawing upon feedback from the verifier LLM, until the verifier LLM either approves a candidate plan as correct or the number of iterations surpasses a predefined threshold. We present an empirical evaluation of (i) the effect of self-critiquing on the plan generation performance of the overall LLM+LLM system (ii) the performance of the verifier LLM in comparison to the ground-truth verification and finally (iii) the influence of varying feedback levels while critiquing the LLM's generation on the overall system performance. For our study, we use GPT-4 \cite{openai2023gpt4} as both the generator and verifier.

Our findings suggest that self-critiquing degrades the plan generation performance compared to when an external, sound verifier is utilized. This decline in performance can be directly attributed to the verifier LLM's subpar results. The verifier LLM yields a significant number of false positives, which can severely undermine the system's reliability. Furthermore, we explored whether the nature of feedback on invalid plans influences plan generation performance. Our results indicate that the type of feedback—whether it's merely binary verification or combined with detailed feedback on the errors of the generated plan—doesn't significantly impact plan generation performance.

Thus, our systematic investigation offers compelling preliminary evidence to question the efficacy of LLMs as verifiers for planning tasks within an iterative, self-critiquing framework. In the rest of the paper, we first present the related work, then the required background before delving into the methodology and the evaluation.

\section{Related Work}
There has been significant interest in investigating the reasoning capabilities of LLMs, spanning from planning \cite{valmeekam2023planning} to logic and arithmetic \cite{dziri2023faith}, and even puzzles \cite{yao2023tree}. As the initial excitement from triumphant anecdotes about LLMs' reasoning capabilities began to wane with systematic studies \cite{valmeekam2023planning, silver2022pddl, dziri2023faith}, researchers proposed that allowing LLMs to verify their own candidate solutions and iterate over this process could enhance their reasoning abilities \cite{shinn2023reflexion, madaan2023self, kim2023language, weng2022large}. Our work systematically investigates the effect of iterative self-critiquing in the context of planning.

There have also been studies that utilize multiple LLMs to generate and verify candidate solutions, either in the form of a debate \cite{du2023improving} or through cross-examination \cite{cohen2023lm}. However, these studies still rely solely on the verification/self-critiquing abilities of the LLMs, an aspect our work critically examines in the context of planning. Our results provide compelling reasons to question the use of LLMs for self-critiquing in planning.

\section{Background}
We specifically are interested in classical planning problems that are represented within the PDDL (Planning Domain and Definition Language) framework \cite{McDermott1998PDDLthePD}. These problem classes consist of a domain, initial state and a goal state. The domain consists of a set of predicates and a set of actions. The state-space of the planning problem is represented with some truth-assignment on the predicates. Every action in domain have a set of preconditions which determine when an action can be applied and a set of effects which determine the modifications to the state after the action is applied. A plan here is a sequence of actions which are present in the domain that when executed in the initial state, satisfy the goal conditions.
\section{Methodology}
\begin{figure}
    \centering
    \includegraphics[width=0.9\linewidth]{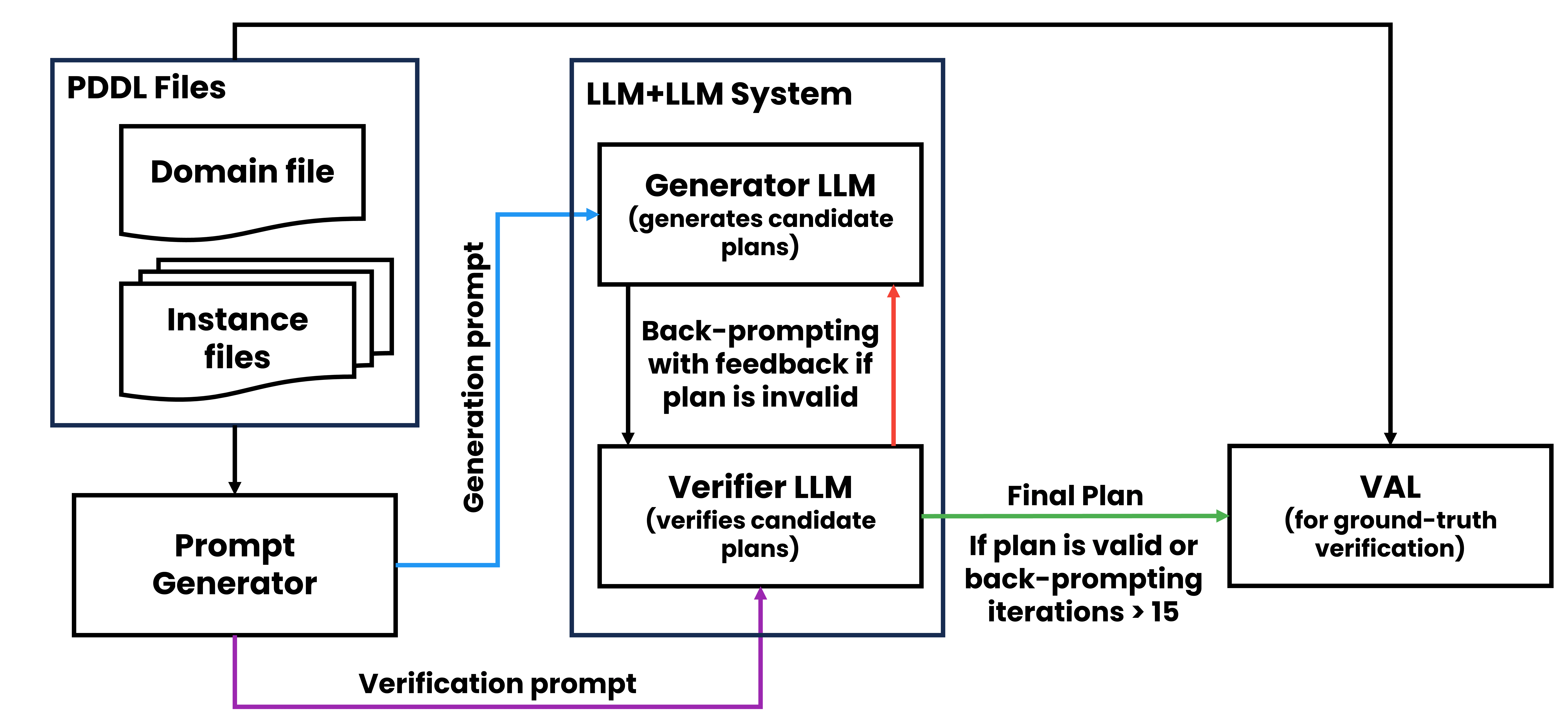}
    \caption{Overall evaluation architecture}
    \label{fig:arch}
\end{figure}
\subsection{The LLM+LLM planning system}
The LLM+LLM planning system (as shown in Figure \ref{fig:arch}) consists of a generator LLM and a verifier LLM. For a given instance, the generator LLM produces a candidate plan, while the verifier LLM determines its correctness. If the plan is found to be incorrect, the verifier provides feedback detailing the reasons for its failure. This feedback is then relayed to the generator LLM, prompting the generation of a new candidate plan. It's worth noting that there are no constraints on the type or format of feedback the verifier LLM produces. The system ceases generation either when the verifier LLM approves the candidate plan as valid or when the number of prompting iterations exceeds a set threshold (for our experiments, this threshold is set at 15 iterations). This method is similar to the backprompting technique described in \cite{valmeekam2023planning}. However, the main distinction lies in the type of verifier employed. In our system, both the verifier and generator are LLMs, whereas the referenced approach utilizes an external sound verifier, VAL \cite{howey2004val}. For all our experiments, GPT-4 serves as the default LLM.
\subsection{Prompt generation}
For the LLM+LLM Planning system described above, we utilize distinct prompts for the generator and verifier LLMs. The prompt generator (as shown in Figure \ref{fig:arch}) utilizes the PDDL domain and instance files to generate the required prompts in natural language. Our prompts are structured similarly to the natural language prompts found in \cite{valmeekam2023planning}. For plan generation, our prompts are one-shot: we begin by presenting the domain description, followed by an example instance (along with its corresponding plan). We then present the query instance. These example instances are randomly selected from our set of instances, and this forms the input for the generator LLM. For the verifier LLM, we adopt a zero-shot approach. Here, we present the domain description, followed by the query instance and its corresponding plan. The verifier LLM is then tasked with verifying the query plan and providing feedback if necessary. As mentioned earlier, we do not restrict the type or format of the feedback for the verifier LLM. Detailed examples of the prompts given to both the generator and verifier LLMs can be found in the Appendix. 
\section{Evaluation and Analysis}
We evaluate our planning system on Blocksworld, a widely recognized common-sense planning domain in AI planning literature \cite{ipc}. We generate 100 random instances for evaluation across various methods. To provide a ground-truth assessment of the final LLM plan's correctness, we employ an external sound verifier, VAL \cite{howey2004val}. For all experiments, GPT-4 \cite{openai2023gpt4} serves as the chosen LLM and was run with a temperature of 0, thereby making it deterministic.

\subsection{Effect of self-critiquing on plan generation}

We assessed the impact of self-critiquing on plan generation by comparing the LLM+LLM backprompting system with two other baselines. The first baseline is the LLM+VAL backprompting system, which mirrors the backprompting method described in \cite{valmeekam2023planning}. In this method, the plan produced by the LLM is validated by an external sound verifier, VAL. If the plan is found lacking, the generator-LLM is reprompted using feedback from VAL. The second baseline involves a generator-LLM without backprompting. Here, the generator LLM receives a single prompt, and the resulting plan is considered final.

As illustrated in Table \ref{tab:accuracy}, the LLM+LLM backprompting approach slightly outperforms the non-backprompting method in terms of accuracy. However, it falls short when compared to the LLM+VAL system. It's worth noting that the marginal improvement over the generator-LLM-only method might not solely be attributed to the LLM verifier. The backprompting itself, which offers the generator LLM multiple opportunities to produce a plan, could be a contributing factor. The subpar performance of the LLM+LLM system, especially when compared to LLM+VAL, can likely be traced back to the substantial number of type-1 errors produced by the LLM verifier. It's evident that incorporating a sound verifier in the backprompting process can significantly enhance overall performance.
\begin{table}[h]
    \centering
    \begin{tabular}{|p{5cm}|c|c|}
        \hline \Tstrut
         Plan Generation Method& Accuracy & Avg. Number of iterations\\ 
         \hline \Tstrut
        LLM+LLM w/ Backprompting (BP) & 55/100 (55\%) & 3.48\\
         \hline \Tstrut
         LLM+VAL w/ BP & 88/100 (88\%) & 4.18\\
         \hline \Tstrut
         Generator LLM only w/o BP & 40/100 (40\%) & 1.00\\
         \hline
    \end{tabular}
    \caption{Comparison between various plan generation methods on the Blocksworld domain.}
    \label{tab:accuracy}
\end{table}

\subsection{Analysis on the self-critique verifier}
We base our evaluation of the verifier LLM on its binary verification (i.e., determining whether the plan is valid or not) of the final plan produced by the LLM+LLM system. It's important to note that the system halts either when the verifier LLM considers the plan valid or when the number of iterations surpasses 15. We compare the LLM verifier's output with ground truth classifications made using VAL \cite{howey2004val}, a sound verifier. To make the ground truth determination available for each input plan, we separately evaluate that plan using VAL as well.

As illustrated in Table \ref{tab:confusion}, out of the 100 instances, the verifier accurately identifies 61 (or 61\%). However, a deeper examination of the verifier's errors reveals a concerning number of false positives. In this context, a false positive refers to the verifier LLM deeming a generated plan valid when, in fact, it is not. Out of the 100 instances, the verifier LLM produces 54 true positives and 38 false positives (type-1 errors). This indicates that the verifier deemed 38 plans, which were actually invalid, to be valid which can be catastrophic if such a system is deployed in scenarios where correctness is paramount.

\begin{table}[h]
    \centering
    \begin{tabular}{|p{1cm}|P{1.9cm}|P{2cm}|P{2.25cm}|P{2cm}|P{2.2cm}|}
    \hline
    \Tstrut          & Accuracy & True Positive Rate & False Positive Rate & True Negative Rate & False Negative Rate \\ \hline
        \Tstrut Verifier LLM & 61/100 (61\%) & 54/55 (98.2\%)  & \textbf{38/45 (84.45\%)} & 7/45 (15.55\%) & 1/55 (1.8\%) \\ \hline
    \end{tabular}
    \caption{Breakdown of Plan Verification results on Blocksworld domain. The denominators (in aspects other than Accuracy) are ground-truth values based on VAL.}
    \label{tab:confusion}
\end{table}

\subsection{Effect of the levels of feedback on plan generation}

While the use of a sound verifier appears to enhance overall performance, we sought to further investigate the impact of varied levels of feedback on plan generation performance. We assessed the system's performance across four distinct feedback levels:

\begin{enumerate}
    \item No Feedback: At this level, the initial plan generated by the LLM is considered to be final and no feedback is provided to the LLM.
    \item Binary Feedback: This level simply indicates whether the generated plan is valid or not.
    \item Inexecutable Action Feedback: If the plan is invalid and inexecutable, this feedback highlights the first inexecutable action and the unmet preconditions causing the inexecutability. If the plan is executable but fails to meet all goal conditions, the unmet goal conditions are presented. This feedback mirrors what VAL provides.
    \item Open Conditions Feedback: This level treats the plan as a partial-order plan \cite{weld1994introduction} and presents all the actions for which there exists atleast one unmet pre-condition and the corresponding unmet preconditions. Further it also presents the unmet goal conditions.
\end{enumerate}

Table \ref{tab:levels} showcases the LLM's performance when subjected to various levels of feedback (including one with no feedback). Interestingly, the amount of feedback provided to the LLM seems to have minimal influence on its performance improvement. As long as the binary feedback is accurate and the LLM is given ample opportunities to generate a plan, the detailed feedback on invalid plans doesn't appear to significantly enhance the LLM's performance. We have provided examples for each feedback level in the Appendix.
\begin{table}[h]
    \centering
    \begin{tabular}{|p{5.5cm}|P{1.9cm}|P{1.5cm}|}
    \hline \Tstrut
        Levels of feedback & Accuracy & Avg. no of steps \\ \hline \Tstrut
        No feedback & 40/100 (40\%) & 1.00 \\ \hline \Tstrut
        Only binary feedback & 37/50 (74\%) & 5.38 \\ \hline \Tstrut
        Binary + First error feedback (by VAL) & 43/50 (86\%) & 4.18 \\ \hline \Tstrut
        Binary + All errors feedback & 43/50 (86\%) & 4.42 \\ \hline
    \end{tabular}
    \caption{Performance of LLM+VAL system on plan generation with varied levels of feedback.}
    \label{tab:levels}
\end{table}

\section{Conclusion and Future Work}
In this paper, we conducted a systematic investigation into the ability of Large Language Models (LLMs) to critique their own outputs, specifically within the context of classical planning problems. While recent research has been optimistic about LLMs' potential in self-critiquing, especially in iterative settings, our findings present a different perspective.

Our empirical evaluations on Blocksworld, a simple common-sense domain, highlighted the ineffectiveness of self-critiquing in LLMs in the context of planning. We showed that the verifier LLM generates a significant number of false positives which be detrimental to the overall system's reliability, particularly in domains where the correctness of plans is paramount. Interestingly, the nature of feedback, whether binary or detailed, did not have a pronounced impact on plan generation performance, suggesting that the core issue lies in the LLM's binary verification capabilities rather than the granularity of feedback.

In the future, we plan to conduct more extensive experiments with respect to the number of instances, the number of domains and prompting methods (such as chain-of-thought).

 \bibliography{llmplan}
\bibliographystyle{plain}


\end{document}